\documentclass{article}

\usepackage{arxiv}

\usepackage[utf8]{inputenc} 
\usepackage[T1]{fontenc}    
\usepackage{hyperref}       
\usepackage{url}            
\usepackage{booktabs}       
\usepackage{amsfonts}       
\usepackage{nicefrac}       
\usepackage{microtype}      
\usepackage{lipsum}
\usepackage{amsmath}
\usepackage{multirow}
\usepackage{adjustbox}

\usepackage{subcaption}
\usepackage{rotating}
\usepackage{tikz}
\usepackage{lscape}
\usetikzlibrary{positioning}
\usepackage{array,calc}
\newlength\lena \newlength\lenb \newlength\lenc \newlength\lend
\newcolumntype{P}[1]{>{\centering\arraybackslash}p{#1}} 

\title{Semi-supervised GANs to Infer Travel Modes in GPS Trajectories}

\author{
  Ali Yazdizadeh \\
  Transportation Research for Integrated Planning (TRIP) Lab\\
  Concordia University \\
  \texttt{e-mail: ali.yazdizadeh@mail.concordia.ca}\\
   \And
  Zachary Patterson\\
	Transportation Research for Integrated Planning (TRIP) Lab\\
	Concordia University \\
	\texttt{e-mail: zachary.patterson@concordia.ca}\\	
	\And   
  Bilal Farooq \\
  Laboratory of Innovations in Transportation (LiTrans)\\
  Ryerson University\\
  Toronto, Canada \\
  \texttt{bilal.farooq@ryerson.ca} \\
}

\begin{document}
\maketitle

\begin{abstract}
Semi-supervised Generative Adversarial Networks (GANs) are developed in the context of travel mode inference with uni-dimensional smartphone trajectory data. We use data from a large-scale smartphone travel survey in Montreal, Canada. We convert GPS trajectories into fixed-sized segments with five channels (variables). We develop different GANs architectures and compare their prediction results with Convolutional Neural Networks (CNNs). The best semi-supervised GANs model led to a prediction accuracy of 83.4\%, while the best CNN model was able to achieve the prediction accuracy of 81.3\%. The results compare favorably with previous studies, especially when taking the large-scale real-world nature of the dataset into account.
\end{abstract}

\keywords{Generative adversarial networks, convolutional neural networks, GPS trajectories, mode inference, smartphone household travel survey }



\section{Introduction}
Inferring travel mode from GPS trajectories is one of the important steps in deriving trip information from smartphone-based travel surveys. Various approaches in the literature have been used to infer mode from GPS traces, such as rule-based classifiers \cite{stopher2009travel}, machine learning 
models \cite{gonzalez2010automating}, deep learning \cite{dabiri2018inferring}. With respect to machine/deep learning methods, there are three types of learning approaches: supervised, semi-supervised and unsupervised learning. In supervised learning, a set of labeled training data is used to infer a function that maps an input to an output based on input-output pairs. Semi-supervised learning is also a supervised learning class but where only a subset of training data is labeled. Semi-supervised learning methods can still take advantage of unlabeled data for training. In unsupervised learning, all training data are unlabeled. Unsupervised learning infers a function that describes the structure of data and groups them. Most of the literature concerning machine learning and information inference from GPS and smartphone data has used supervised learning approaches. Moreover, deep learning (supervised) algorithms typically show their best performance on extremely large labeled datasets \cite{goodfellow2016nips}. One of the main challenges of these methods is thus access to such datasets. Semi-supervised learning is one flexible strategy to reduce the required number of labeled examples by studying large unlabeled data sets, which are easier to obtain \cite{goodfellow2016nips}.

Recently, Generative Adversarial Networks (GANs) \cite{goodfellow2014generative} have shown promising results in image and language processing. In the GANs framework, a generative model is set against an adversary, or a ``discriminative'' model, that learns to distinguish between the observations produced by a ``generative model'' and real data observations \cite{goodfellow2014generative}. While the GANs framework has mainly been used for generating samples, it can be trained for semi-supervised learning, where the labels for a considerable part of examples are missing. One example is CatGANs \cite{springenberg2015unsupervised}, which successfully trained a discriminator classifier on unlabeled and partially labeled data. 

This study investigates the application of semi-supervised GANs to infer travel mode from the GPS traces of travellers. The paper is organized as follows: a background section describes the literature related to transportation mode inference and generative adversarial networks. Next, the methodology section introduces the data processing steps. Afterwards, the results section describes the developed models and their prediction results. Finally, we present our conclusions and future directions.

\section{Background}\label{litreview}
This section reviews previous research related to machine learning used in mode detection from smartphone data. We also briefly describe the GANs and semi-supervised GANs frameworks.

\subsection{Mode Detection and Machine Learning}
Mode detection methods have been applied on various data sources including raw GPS trajectories \cite{yazdizadeh2018automated, dabiri2018inferring, rezaie2017semi}, accelerometer data from smartphones \cite{eftekhari2016inference}, and Wi-Fi signals \cite{Kalatian2018}. 

Endo et al. \cite{endo2016deep} attempted to use deep neural networks to automatically extract high-level features. They introduced an innovative idea to convert a raw GPS trajectory into a 2-D image structure as the input into the deep neural network model. Instead of RGB (Red, Green, Blue) values of an image pixel, they equivalently consider the duration time that a user stays in the location of the pixel. They deployed traditional classifiers, such as logistic regression and decision tree, to predict transportation mode. Their best models were able to detect the travel mode with prediction accuracy of 67.9\%.

Assemi et al. \cite{assemi2016developing} developed a nested logit model to infer travel mode from smartphone travel surveys, using eight attributes. Their model can predict travel mode on GPS data gathered in Australia with 79.3\% of accuracy, which does not include any data pre-processing step. They have also reported an accuracy of 97\% based on GPS data gathered in New Zealand. However, they have mentioned that the high accuracy has been gained after undergoing an excessive data pre-processing step, including data cleaning.
Wang et al. \cite{wang2017detecting} used CNN models for deep feature learning. They categorized the attributes into two classes: the point-level features and the trajectory-level features. They selected speed, heading change, time interval, and distance as the point-level handcrafted features. They considered average speed, variance of speed, heading change rate, stop rate, and speed change rate as the trajectory-level handcrafted features. Finally, they have feed both type of features into a a deep neural network classifier.

More recently, Dabiri and Heaslip \cite{dabiri2018inferring} used Convolutional Neural Network (CNN) to train a mode detection model with an accuracy of 79.8\%. They implemented different CNN models on GPS trajectories, and finally combined the output of the CNN models via an ensemble method. Their ensemble library was made up of seven CNN models from which they took the average of softmax class probabilities, predicted by each CNN model, to generate the transportation label posteriors. 


\subsection{Generative Adversarial Networks}
GANs were introduced primarily as unsupervised learners that benefit from setting up a supervised learning framework. However, besides unsupervised learning, GANs are also able to undertake semi-supervised learning. Semi-supervised GANs have been implemented via three different approaches in the literature. Class-conditional GANs were first introduced by Mirza and Osindero \cite{mirza2014conditional}. As Mirza and Osindero \cite{mirza2014conditional} have explained, the unconditioned (traditional) generator in GANs has no control on the class labels of the data being generated. Mirza and Osindero \cite{mirza2014conditional}, proposed a model able to direct the generating process by conditioning the generator on additional information 
\cite{mirza2014conditional}. They explain how GANs are able to do multi-class labelling by conditioning the generator on class labels. Their proposed model can generate samples for each class. For example, in image classification problems, you can ask Conditional GANs for a ``horse'' class, and it will produce a picture of a horse. 

The second approach for semi-supervised GANs enables GANs to predict K different output classes. In this approach, instead of predicting one ``real'' class and one ``fake'' class, the GANs predicts K different real classes. Fake (generated) data should result in the GANs being not confident about which class to output. This approach was first developed by Springenberg \cite{springenberg2015unsupervised} in the CatGAN mode.

\begin{figure}[!t]
	\centering
	\includegraphics[keepaspectratio,width=1.02\linewidth]{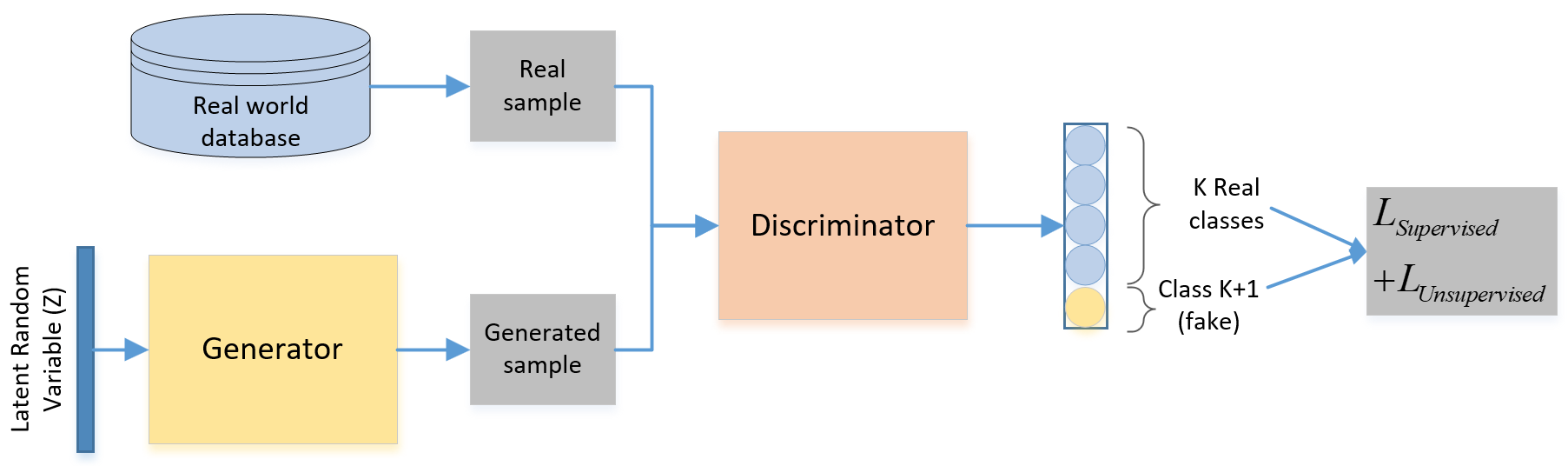}
	\caption{The Framework of Semi-supervised Generative Adversarial Network.
		\label{fig:semi_GANs}}
\end{figure}

In the third approach, the GANs model outputs K+1 different classes. This semi-supervised GANs approach was first introduced by Salimans et al. \cite{salimans2016improved} and Odenda \cite{odena2016semi}. Salimans et al. \cite{salimans2016improved} added samples from the GANs generator G to their data set, as a new ``generated" class, and then used a standard classifier to do the semi-supervised learning. So, if we consider a standard classifier that classifies a data point x into one of K possible classes, the semi-supervised classifier will classify the data point into K+1 classes, with the K+1th class containing the observations ``generated" by a GAN \cite{salimans2016improved}, as shown in Figure \ref{fig:semi_GANs}. 

While GANs is mainly used as a generative model to generate samples through labeling the data into ``fake'' and ``real'' \cite{goodfellow2016nips}, some studies have extended GANs for multi-class labeling. However, almost all implementations of GANs models have been conducted in image processing or speech recognition studies. Analyzing GPS trajectories with semi-supervised GANs models to detect the travel mode has not been applied in the literature yet. 

\section{Research Design and Methodology}\label{methodology}
This section describes the data gathered from the smartphone travel survey, the architectures of semi-supervised GANs models that are developed. Moreover, a data preparation section describes the pre-processing steps implemented on the trajectory data.

\subsection{Data} \label{data}
This study used data gathered by the MTL Trajet \cite{patterson2017mtl} smartphone Travel Survey App in the fall of 2016. The entire dataset comprises more than 33 million GPS points from over 8,000 respondents. To identify trips and segments, the trip-breaking algorithm, which is a rule-based algorithm, developed in Patterson \& Fitzsimmons \cite{patterson2016datamobile} was used. The algorithm recognizes segments whenever it detects a 3-min gaps in data while checking speed and the public transit network locational data. After implementing the trip-breaking algorithm on the GPS points, 623,718 trips were detected, among which 102,904 trips were validated by respondents. Afterwards, we applied several pre-processing and segmentation steps explained in the following.

\subsubsection{Data Preparation}
The input layer to the CNN or DCGAN models need to be of fixed size. However, the number of GPS points along a trip are not equal for all the trips in the MTL Trajet dataset. Hence, we split the detected trips into fixed size segments. We examined the average (70 points) and median (120 points) number of points for the length of the fixed size segments. Following testing both 70 and 120 point segments, we observed stronger performance of neural network models on seventy-point segments. We padded with zero when the segment size was less than seventy points.

With respect to the labelling method, the smartphone app prompted travellers to validate their travel mode whenever a stop was detected throughout their movements during the day. In addition, respondents were asked to declare their travel mode to main destinations by answering a questionnaire wheninstalling the MTL Trajet application. In particular, respondents were asked to reveal the location of home, school and work and home, along with the transportation mode(s) used to travel to these locations. We considered only the validated trips of those users who stated that they used only one mode option to make a trip between home and work/school. The final dataset thereby consists of 3845, 8515, 7415 and 15275 walk, bike, transit and car segments, respectively. 

\begin{figure}[h]
	\centering
	\includegraphics[keepaspectratio,width=1\linewidth]{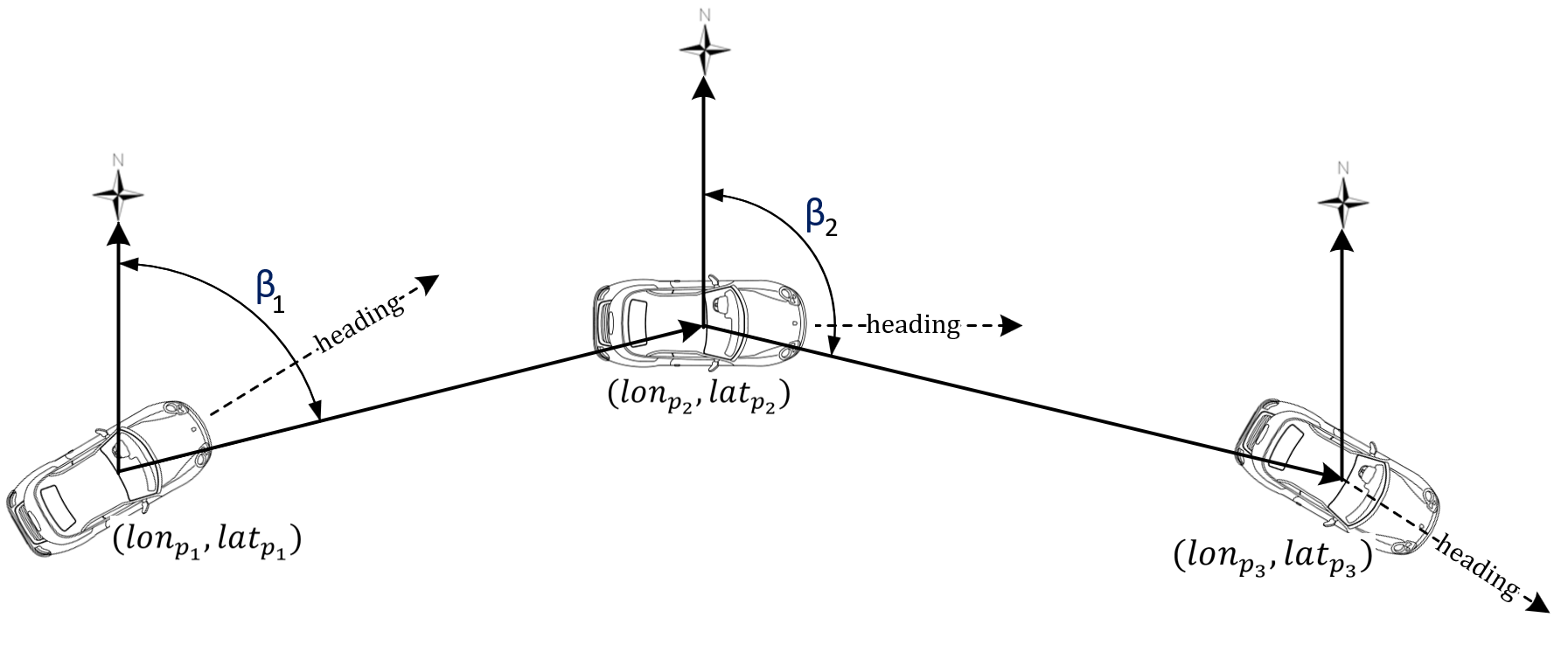}
	\caption{Bearing ($\beta_1 and \beta_2$) and Heading of a Moving Vehicle.
		\label{fig:bearing}}
\end{figure}

We calculated five characteristics for each point along a segment: ``distance to previous point'', ``speed'', ``acceleration'', ``jerk'', and ``bearing rate'', which have been used in several studies \cite{dabiri2018inferring,zaki2014use,girshick2015deformable}. These characteristics are shown as channels of each segment in Figure \ref{fig:70-points-5channels}. Afterwards, these segments were fed as input to the discriminator of the semi-supervised DCGANs model. Jerk describes the rate of change of acceleration. Bearing, as shown in Figure \ref{fig:bearing}, is defined as the angle between magnetic north and the direction of a point to its consecutive next point.

We can obtain the bearing by applying the following equations on any two consecutive GPS points, i.e. $p_1$,and $p_2$: \cite{dabiri2018inferring}:

\begin{figure}[b]
	\centering
	\includegraphics[keepaspectratio,width=0.9\linewidth]{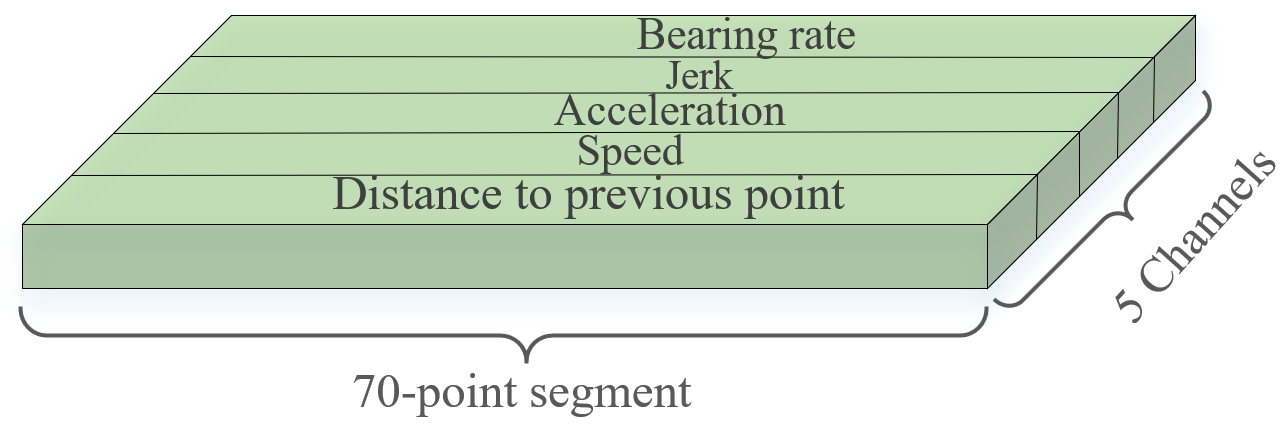}
	\caption{A 70-point Segment with 5 Channels.
		\label{fig:70-points-5channels}}
\end{figure}

\begin{flalign}
\beta_1 &=\arctan (X,Y)&&
\end{flalign}
where:
\begin{flalign}
X&=\cos(lat_{p_1})*\sin(lat_{p_2})-\sin(lat_{p_1})*&&\\\nonumber
&\cos(lat_{p_2})*\cos(lon_{p_2}-lon_{p_1})&&\\
Y&=\sin(lon_{p_2}-lon_{p_1})*\cos(lat_{p_2})
\end{flalign}

The $lon$ and $lat$ are the longitude and latitude of GPS points along a segment, and are in radians. The formula for calculating the bearing rate is as follows \cite{dabiri2018inferring}:

\begin{flalign}
Bearing\, Rate &=\mid \beta_2 - \beta_1 \mid&&
\end{flalign}

Obviously from this formula, the bearing rate is equal to the absolute change in the bearing of two consecutive points. Also, it is necessary to have at least three consecutive points to measure the bearing rate (as demonstrated in Figure\ref{fig:bearing}), because the bearing of each point is calculated according to its following point. 

\subsection{Details OF GANs Training}
The most well-known GANs architecture is the Deep Convolutional GAN (DCGAN) \cite{radford2015unsupervised}. Most GANs use at least some of the architectural innovations proposed in the DCGAN architecture. We trained the semi-supervised DCGANs on the MTL Trajet dataset. 

Hyper-parameter optimization is considerably important in training GANs. As Lucic et al. \cite{lucic2018gans} have mentioned, different GANs algorithms can arrive at similar performance with enough hyper-parameter optimization. we trained models based on hyper-parameters values previously examined in the literature and random values that we think are useful to be tested. We tested different number of layers for discriminator and generator from 2, 3, 4, 12. Also, With respect to the kernel size, different values were tested, from 4,8,16,and 32. The number of filters per each layer of discriminator and generator were tested based on following values: 8, 16, 32, 64, 128, 256, 512.

We trained different models with different batch sizes of 16, 32, 64 and 128. The models with a batch size of 16 demonstrated better results. With respect to the activation function, We examined 'Relu', 'Leaky Relu' and 'tanh' functions and got higher prediction accuracies with Leaky Relu function. 

We applied one-sided label smoothing \cite{salimans2016improved} to the positive labels of the generator. Label smoothing is a technique that replaces the 0 and 1 labels for a classifier with smoothed values like 0.9 or 0.1. This technique has been proven to enhance the performance of neural networks while facing adversarial examples \cite{warde201611}. We used the Adam optimizer for training both generator and discriminator. In addition, gradient clipping was used to keep the training procedure in steady state. Gradient clipping prevents the norm of the gradient from exceeding a given value \cite{goodfellow2016deep}. In the next section the results of best trained GANs architecture for detecting GPS trajectories have been presented. Moreover, we trained different CNN architectures and presented the best models as the baseline models to be compared with GANs.

\subsection{The Model Architecture}
Table~\ref{table:architecture} shows the architecture of the different CNN and DCGAN models developed in this paper. Models A, B and C, in Table~\ref{table:architecture}, are the CNN models and Models E and F are the semi-supervised GANs models.

\begin{table*}[]
	\centering
	\caption{Architecture of Convolution Neural Networks and Semi-supervised GANS}
	\label{table:architecture}
	\scriptsize
	\begin{adjustbox}{width=1.0\textwidth,totalheight=1\textheight,keepaspectratio}
		\begin{tabular}{ccccclcc}
			\hline
			Model A                                                      & Model B                                                      & Model C                                                      & \multicolumn{2}{c}{Model D}                                                            &  & \multicolumn{2}{c}{Model E}                                                            \\ \hline
			\multirow{2}{*}{CNN}                                         & \multirow{2}{*}{CNN}                                         & \multirow{2}{*}{CNN}                                         & \multicolumn{2}{c}{Semi-supervised GANS}                                               &  & \multicolumn{2}{c}{Semi-supervised GANS}                                               \\ \cline{4-5} \cline{7-8} 
			&                                                              &                                                              & Generator                                                      & Discriminator         &  & Generator                                                      & Discriminator         \\ \cline{1-5} \cline{7-8} 
			Input $[70 \times 5]$                                        & Input $[70 \times 5]$                                        & Input $[70 \times 5]$                                        & Z {[}100{]}                                                    & Input $[70 \times 5]$ &  & Z {[}100{]}                                                    & Input $[70 \times 5]$ \\ \hline
			\begin{tabular}[c]{@{}c@{}}CONV8-32\\ MAXPOOL8\end{tabular}  & \begin{tabular}[c]{@{}c@{}}CONV8-128\\ MAXPOOL8\end{tabular} & \begin{tabular}[c]{@{}c@{}}CONV8-96\\ MAXPOOL8\end{tabular}  & \begin{tabular}[c]{@{}c@{}}Projection\&\\ reshape\end{tabular} & CONV8-32              &  & \begin{tabular}[c]{@{}c@{}}Projection\&\\ reshape\end{tabular} & CONV8-128             \\ \hline
			\begin{tabular}[c]{@{}c@{}}CONV8-64\\ MAXPOOL8\end{tabular}  & \begin{tabular}[c]{@{}c@{}}CONV8-256\\ MAXPOOL8\end{tabular} & \begin{tabular}[c]{@{}c@{}}CONV8-256\\ MAXPOOL8\end{tabular} & FS-CONV8-128                                                   & CONV8-64              &  & FS-CONV8-512                                                   & CONV8-256             \\ \hline
			\begin{tabular}[c]{@{}c@{}}CONV8-128\\ MAXPOOL8\end{tabular} & \begin{tabular}[c]{@{}c@{}}CONV8-512\\ MAXPOOL8\end{tabular} & \begin{tabular}[c]{@{}c@{}}CONV8-384\\ MAXPOOL8\end{tabular} & FS-CONV8-64                                                    & CONV8-128             &  & FS-CONV8-256                                                   & CONV8-512             \\ \hline
			FC                                                           & FC                                                           & \begin{tabular}[c]{@{}c@{}}CONV8-384\\ MAXPOOL8\end{tabular} & FS-CONV8-32                                                    & FC                    &  & FS-CONV8-128                                                   & FC                    \\ \hline
			&                                                              & \begin{tabular}[c]{@{}c@{}}CONV8-256\\ MAXPOOL8\end{tabular} & FS-CONV8-5                                                     &                       &  & FS-CONV8-5                                                     &                       \\ \hline
			&                                                              & FC                                                           &                                                                &                       &  &                                                                &                       \\ \hline
			\multicolumn{8}{l}{CONV: convolution operation, MAXPOOL: Max pooling operation, FS-CONV: Fractionally-strided convolution, FC: Fully-connected}                                                                                                                                                                                                                                
		\end{tabular}
	\end{adjustbox}
\end{table*}

\begin{table}[]
	\centering
	\begin{tabular}{lllll}
		\hline
		Model A & Model B & Model C & Model D & Model E \\ \hline
		76.2\%  & 78.4\%  & 81.3\%  & 81.6\%  & 83.2\%  \\ \hline
	\end{tabular}
	\caption{Prediction Accuracy Rate of Different Models}
	\label{table:Prediciton_Accuracy}
\end{table}

\begin{figure*}[ht]
	\centering
	\begin{subfigure}{\linewidth}
		\centering
		\includegraphics[keepaspectratio,width=0.9\linewidth]{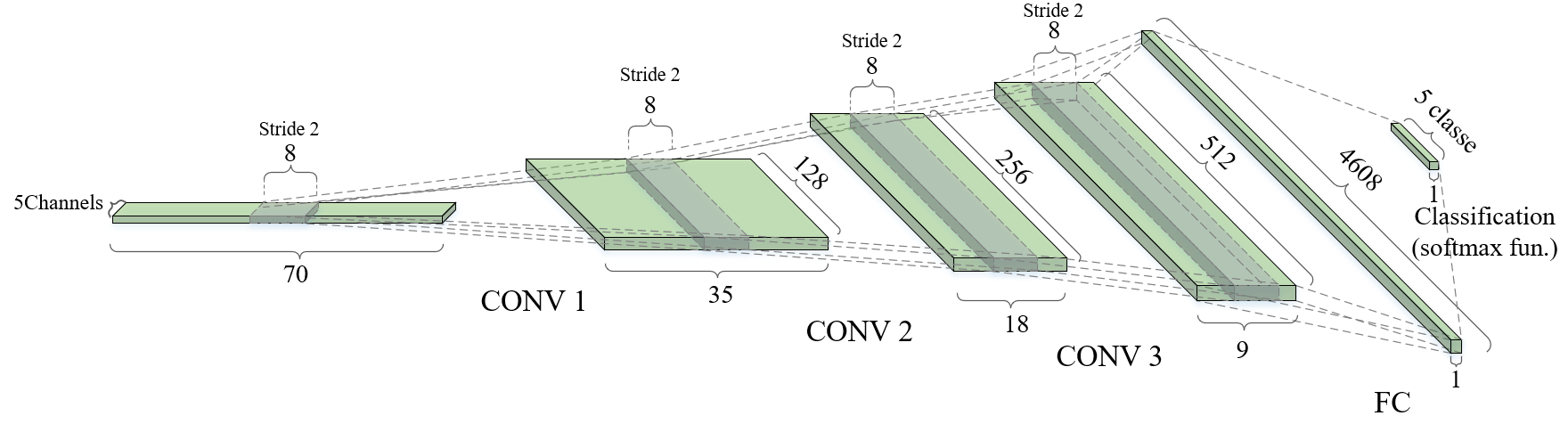}
		\caption{Discriminator Architecture in Model E}
		\label{fig:discriminator-architecture}
	\end{subfigure}
	\hfill
	\begin{subfigure}{\linewidth}
		\centering
		\includegraphics[keepaspectratio,width=0.9\linewidth]{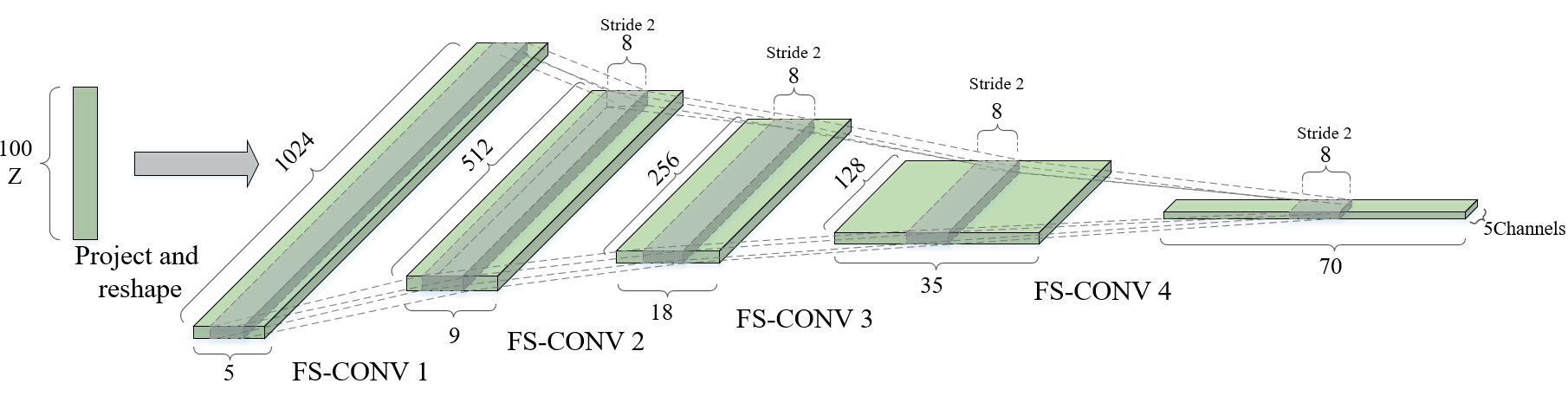}
		\caption{Generator Architecture in Model E}
		\label{fig:generator-architecture}
	\end{subfigure}  
	
	\label{fig:GANS_Architecture}
\end{figure*}

The generator and discriminator architecture of Model E are illustrated in Figure~\ref{fig:discriminator-architecture}. We used convolutional layers for training the discriminator in the DCGANs framework. There are usually three types of layers in a typical CNN architecture: convolutional, pooling and fully-connected layers. We did not use any pooling layer in the DCGAN discriminator as suggested by Radford et al.~\cite{radford2015unsupervised}. Our discriminator, in Model E, begins with segments of size $70\times5$ that contains the 70 GPS points and 5 channels. Afterwards, three convolution operations with kernel size of 8 and stride of 2 are used. The number of kernels in each layer is equal to 128, 256, and 512, respectively. The first convolution operation converts the input segment into a $35\times128$ output. The output of the second and third convolutions are of size $18\times256$ and $9\times512$, respectively. Afterwards, a fully connected operation is applied to the output of the third convolution layer and produces a fully-connected layer of size $1\times4608$. The output of the last fully-connected layer is fed to a 5-way softmax which produces a distribution over the 4 modes of transportation plus the fake class, in other words the discriminator has K+1 output units corresponding to [Class-1, Class-2,..., Class-K, Fake].

Model C is an CNN model based of the architecture architecture developed by Krizhevsky~\cite{krizhevsky2012imagenet} for convolutional neural networks. We used the same number of kernels of CNN model corresponding to the values suggested by the aforementioned study.

The LeakyReLU activation function \cite{maas2013rectifier} is applied to the output of every convolutional layer (the activation functions on hidden layers are not shown in Figure~\ref{fig:discriminator-architecture}). Also, we applied batch normalization \cite{ioffe2015batch} after each convolutional layer and dropout with keep probability of 0.5.  

The generator architecture is shown in Figure~\ref{fig:generator-architecture}. The generator begins with the input z of size 100 which is sampled from a random uniform distribution. The input z is reshaped and projected into a $1024\times5$ output. After, there are four fractionally-strided convolution layers. The output of the final layer is a 70-point segment with 5 channels which generates the fake samples. Finally, the generated fake samples, along with the 70-point segments from the MTL Trajet dataset, are fed to the discriminator as the input layer.

\section{Results}
The results are shown in Table~\ref{table:Prediciton_Accuracy}. The CNN models A, B and C can achieve the prediction accuracy of 76.2\%, 78.4\% 81.3\%, respectively. The semi-supervised DCGANs models D and E predict the travel mode with 81.6\% and 83.2\% accuracy. In total, the developed DCGAN models were able to predict the travel mode with higher accuracy than the CNN models. All the results were calculated using five fold cross-validation methods on MTL Trajet dataset. 

One of the major issues related to generative models is evaluating the generated samples \cite{salimans2016improved}. While in the image processing context the generated images can be evaluated visually, it is not the case for generated trajectory segments. Finding proper evaluation metrics for generated trajectories, with different architectures or hyper-parameters values, should be the focus of future studies. One of the disadvantages of the GANs models is their high computational time and memory requirements. For example, it took around 26 hours to train Model E (with mini-batch size of 16) on two NVIDIA P100 Pascal GPUs and 128 GiB of available memory.

\begin{figure}[!h]
	\centering
	\includegraphics[keepaspectratio,width=1\linewidth]{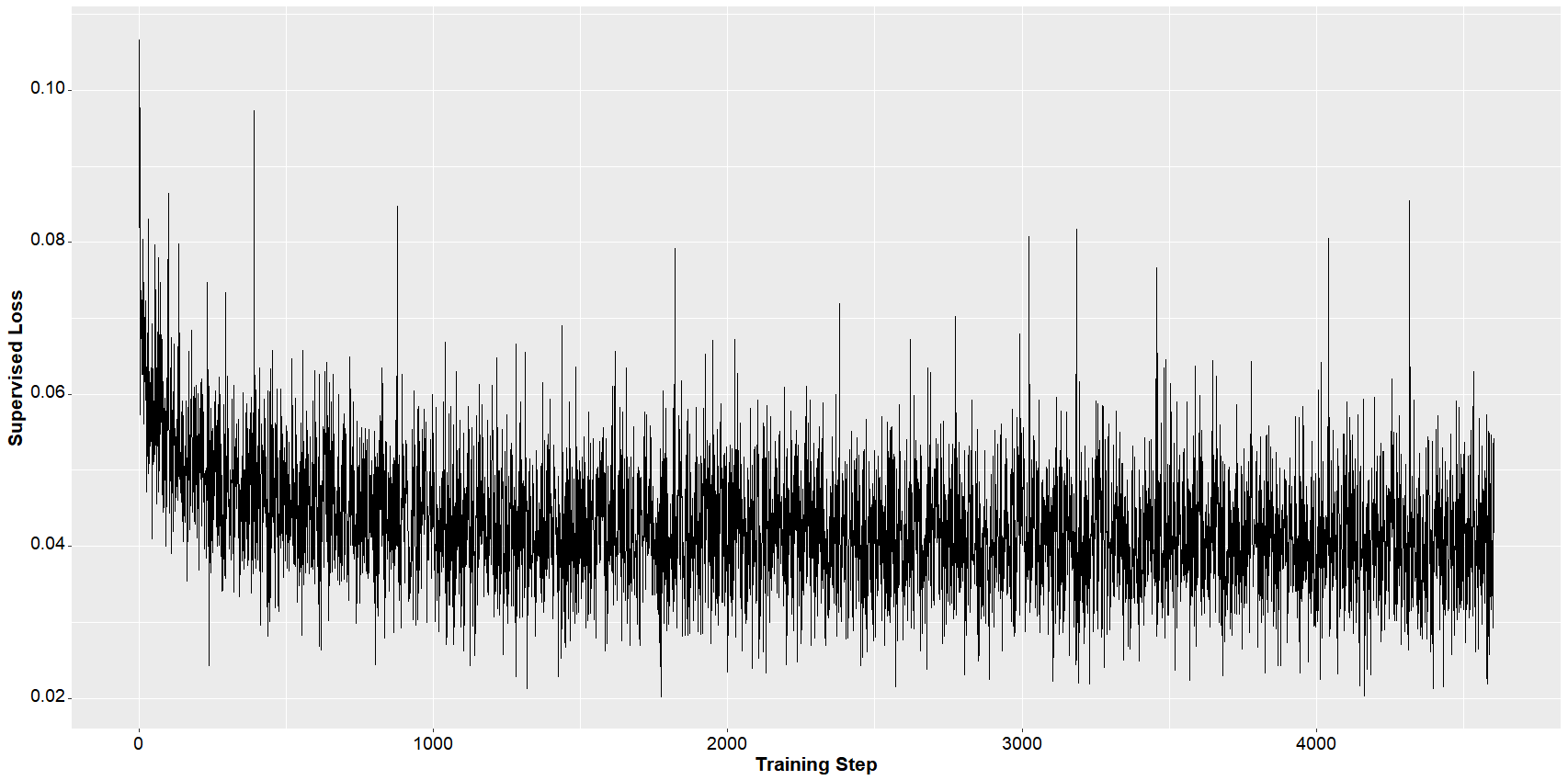}
	\caption{Supervised Loss of DCGANs Model for Different Number of Training Steps.
		\label{fig:Supervised loss}}
\end{figure}

\begin{figure}[!h]
	\centering
	\includegraphics[keepaspectratio,width=1\linewidth]{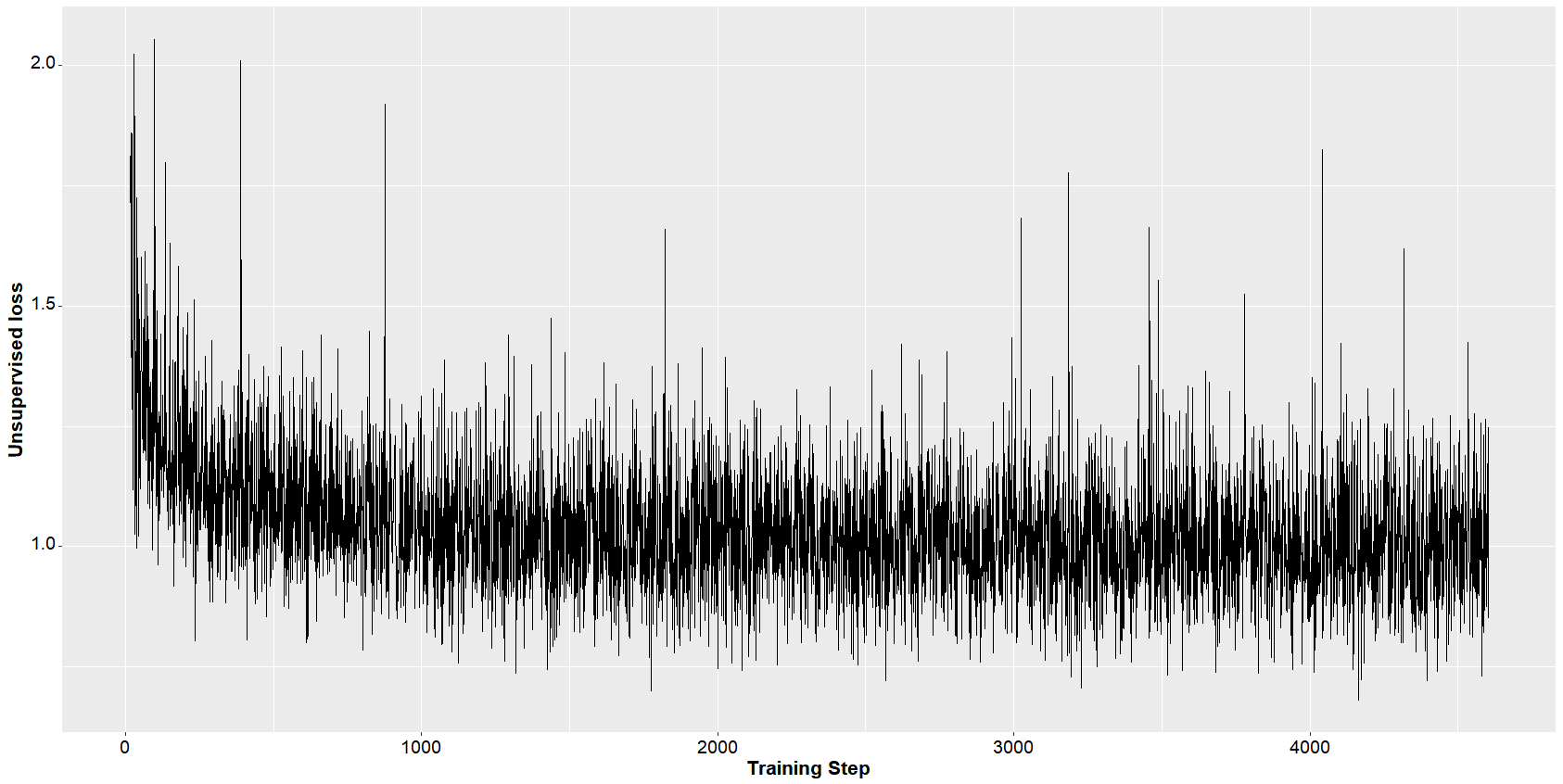}
	\caption{Unsupervised Loss of DCGANs Model for Different Number of Training Steps.
		\label{fig:Unsupervised loss}}
\end{figure}

\begin{figure}[!h]
	\centering
	\includegraphics[keepaspectratio,width=1\linewidth]{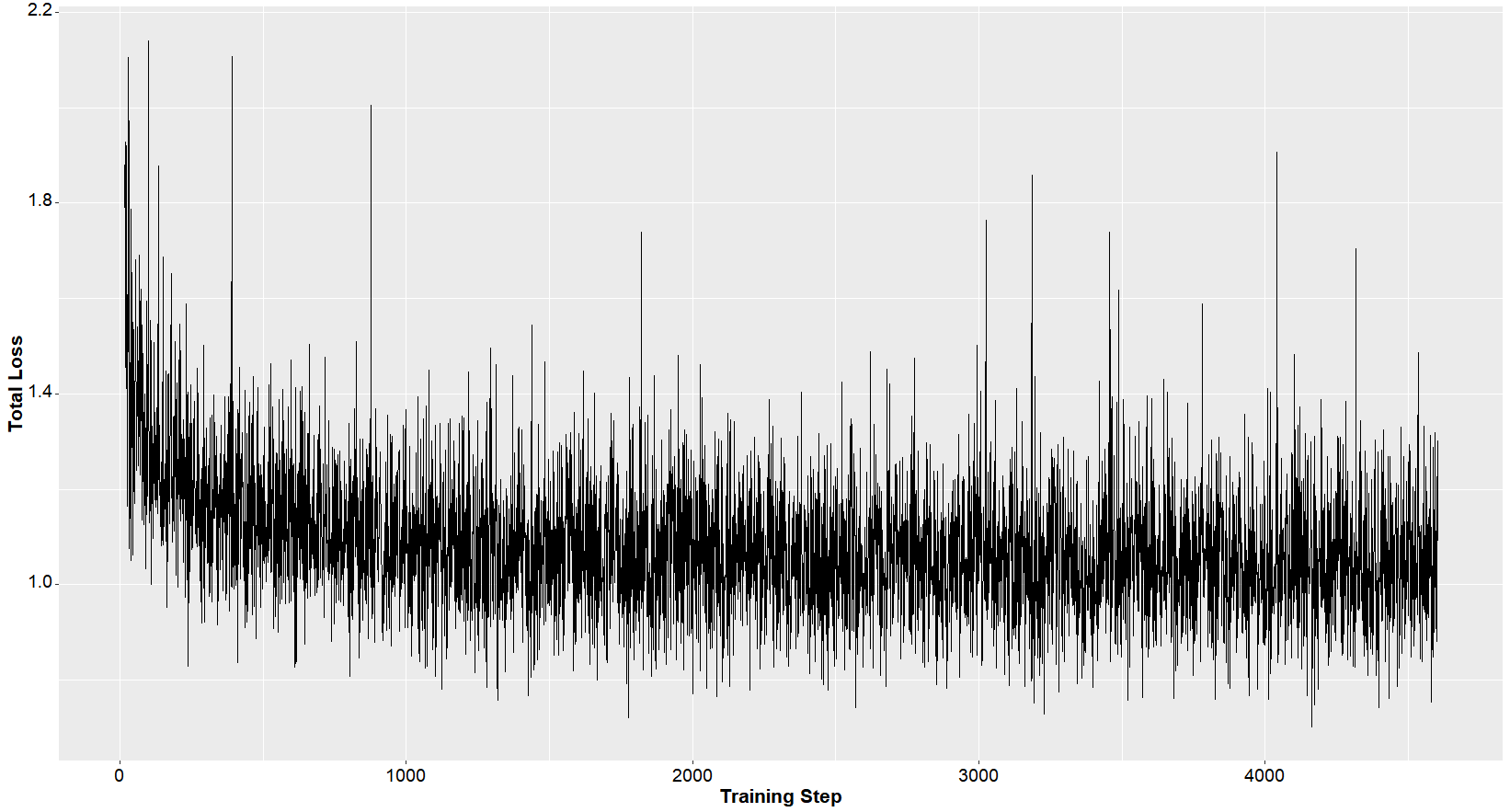}
	\caption{Total Loss of DCGANs Model for Different Number of Training Steps.
		\label{fig:Total loss}}
\end{figure}

The supervised, unsupervised, and total loss for different numbers of training steps are shown in Figures~\ref{fig:Supervised loss}-\ref{fig:Total loss}. The supervised loss is the negative log probability of labels, given that the data come from the real dataset.
The unsupervised loss is composed of two terms: a) the negative log of 1 minus the probability of fake labels given that the data is real, b) the negative log probability of fake labels given that the data is generated (being fake). The unsupervised loss function introduces an interaction between the generator and discriminator (i.e. the classifier), which is indeed the GANs game value \cite{salimans2016improved}. Obviously, the values of unsupervised loss are higher than those of the supervised loss, due to the fact that the generator, i.e. G(z), produces samples by transforming the random noise vectors, i.e. vector z in Figure\ref{fig:generator-architecture}, to the real data distribution \cite{salimans2016improved}.

\subsection{Comparison with previous studies}
While comparing the prediction accuracy rates of different studies, several considerations should be taken into account, such as differences in the quality of data across studies, the number of output categories (classes), and sample sizes. Moreover, the methods of validating the labels of observations in a dataset may affect data quality and the prediction accuracy rate of the developed models. The validation of MTL Trajet data was carried out without any recall or surveyor-intervened validation process. Indeed, the respondents validated their travel mode through an in-app questionnaire. Although such a method of validation places less burden on respondents, it may reduce the quality of the labelled data. Furthermore, the MTL Trajet data comprises primarily GPS data (i.e. no accelerometer data) from user smartphones, to cut down on battery consumption.

That said, and taking into account these considerations, our results are on par with previously obtained findings in the mode detection literature. Dabiri and Heaslip \cite{dabiri2018inferring} have arrived at a test accuracy of 79.8\% for their best convolutional neural network model. Also, the ensemble of their best model was able to reach an accuracy as high as 84.8\%. Zheng et al. \cite{zheng2008understanding} have developed a decision tree model with an overall accuracy of 76.2\%. The aforementioned studies \cite{dabiri2018inferring,zheng2008understanding} have used only GPS data from smartphones, similar to the data gathered by MTL Trajet. The results of our best semi-supervised DCGANs model outperforms the findings of both indicated studies, that is 83.2\% versus 79.8\% and 76.2\% of prediction accuracy, respectively. Apart from the aforementioned differences between the databases and different modeling approaches among the studies, we think that DCGANs model improves the prediction accuracy of discriminator by generating more samples and enlarging the original MTL Trajet dataset.

Bantis and Haworth \cite{bantis2017you} built several classifiers to infer travel mode using GPS and accelerometer data as well as user socio-demographics. Their hierarchical dynamic Bayesian network can reach an accuracy of 90\%. However, their model was built on a training data containing trips from only 5 individuals. Apart from the different modeling approaches between our study and theirs, the richness of the accelerometer data together with the different sizes of datasets may give an explanation to differences in the prediction accuracy. 

\section{Conclusion and Future Work}
Generative adversarial networks are the state-of-the-art in deep learning. The field is growing fast due to its ability to generate new samples by learning the structure and distribution underlying real-world samples. It has proven its ability in unsupervised and semi-supervised image and text recognition. This study developed semi-supervised generative adversarial networks to infer transportation mode from GPS trajectories. 

The developed semi-supervised DCGANs models share the same architectural innovations used in the image recognition literature, that we have now shown can be used in travel information inference from smartphone travel survey data. Generative models have the advantage of increasing the prediction accuracy of classifiers (as seen here) without increasing the number of labeled samples. The semi-supervised DCGANs model in this article prove their superiority over the best developed CNN models. However, in general GANs models have gained a reputation for being difficult to train. One of the major challenges in training GANs models is finding optimal hyper-parameter values and long computation times. As a result, computation/accuracy trade-offs may need to be made when evaluating the use of such models in the future.

In future work the framework developed in this study could be extended along the following dimensions. Apart from the semi-supervised DCGANs developed in this study, there are other types of semi-supervised GANs such as Conditional GANs \cite{mirza2014conditional}, that enable the GANs generator to generate labeled samples, that could be tested. Examining different distributions to sample $z$ from while training the generator to see how it affects the performance of the discriminator. GANs require more training time and memory resources compared to CNN or other machine learning models. The most time-consuming part of GANs is the gradient descent calculations applied simultaneously on both generator and discriminator. Developing methods that need fewer gradient descent steps could significantly decrease the computation time of GANs. The exploration of methods of inferring multiple modes for trips based on smartphone travel survey data. Creating an ensemble of DCGANs models to achieve higher prediction accuracies.

While the prediction accuracies of the models in this paper are not extremely high, they are good relative to other comparable studies in the literature. As well, the fact that the dataset used was from a large-scale real-word study likely introduces a great deal more variability than controlled, small-scale studies. Finally, the purpose of this paper was to demonstrate the use of DCGANS developed primarily for image processing in the context of mode inference with 1-D smartphone trajectory data. Future work will allow exploration of better performing models either with more channels and/or improved architectures.

\bibliographystyle{abbrv}
\bibliography{references.bib}
\end{document}